\documentclass[fleqn,10pt,lineno]{wlpeerj}
\usepackage{listings}
\usepackage{tabularx}
\usepackage{array}
\newcolumntype{P}[1]{>{\hspace{0pt}}p{#1}}

\title{Towards FAIR protocols and workflows: The OpenPREDICT case study}

\author[1]{Remzi Celebi}
\author[2]{Joao Rebelo Moreira}
\author[3]{Ahmed A. Hassan}
\author[4]{Sandeep Ayyar}
\author[5]{Lars Ridder}
\author[2]{Tobias Kuhn}
\author[1]{Michel Dumontier}
\affil[1]{Institute of Data Science,  Maastricht University, Maastricht, Netherlands}
\affil[2]{Computer Science, VU University Amsterdam, Amsterdam, Netherlands}
\affil[3]{Pharmacology \& Personalised
Medicine, Maastricht University, Maastricht, Netherlands}
\affil[4]{Medical Informatics, Stanford University, Palo Alto, California, USA}
\affil[5]{Netherlands eScience Center, Amsterdam, Netherlands}
\corrauthor[1,2]{Remzi Celebi, Joao Rebelo Moreira}{remzi.celebi@maastrichtuniversity.nl, j.l.rebelomoreira@vu.nl}

\begin{abstract}
It is essential for the advancement of science that scientists and researchers share, reuse and reproduce workflows and protocols used by others. The FAIR principles are a set of guidelines that aim to maximize the value and usefulness of research data, and emphasize a number of important points regarding the means by which digital objects are found and reused by others. The question of  how to apply these principles not just to the static input and output data but also to the dynamic workflows and protocols that consume and produce them is still under debate and poses a number of challenges. In this paper we describe our inclusive and overarching approach to apply the FAIR principles to workflows and protocols and demonstrate its benefits. We apply and evaluate our approach on a case study that consists of making the PREDICT workflow, a highly cited drug repurposing workflow, open and FAIR. This includes FAIRification of the involved datasets, as well as applying semantic technologies to represent and store data about the detailed versions of the general protocol, of the concrete workflow instructions, and of their execution traces. A semantic model was proposed to better address these specific requirements and were evaluated by answering competency questions. This semantic model consists of classes and relations from a number of existing ontologies, including Workflow4ever, PROV, EDAM, and BPMN. This allowed us then to formulate and answer new kinds of competency questions. Our evaluation shows the high degree to which our FAIRified OpenPREDICT workflow now adheres to the FAIR principles and the practicality and usefulness of being able to answer our new competency questions.
\end{abstract}

\begin{document}

\flushbottom
\maketitle
\thispagestyle{empty}

\section*{Introduction}

Reproducible results are one of the main goals of science. A recent survey, however, showed that more than 70\% of researchers have been unsuccessful in trying to reproduce another research experiment and more than 50\% failed to reproduce their own research studies \citep{Baker2016}. Failure to represent reproducible methodology is a major detriment to research.
\vspace{\baselineskip}

The rate of non-reproducibility for pharmacological studies is particularly worrying. This, coupled with the incredible expense (over \$1 billion US) and high rate of failure (90\%) demand altogether new approaches for drug discovery \citep{scannell2012diagnosing}. Therefore, pharmacology makes a very good example and use case for the work we present here. Specifically, we will be looking into drug repositioning, where small molecules approved for one indication are repurposed for a new indication. Drug repositioning is gaining recognition as a safe, effective, and much lower-cost approach to uncover new drug uses \citep{ashburn2004drug,sleigh2010repurposing} . The emergence of public data, fueled by a combination of literature curated knowledge and ‘omics data, that has created exciting opportunities for computational drug repositioning. For instance, gene expression data in repositories such as Gene Expression Omnibus (GEO) enables the analysis of correlation between drug and gene expression – termed the Connectivity Map approach - to find chemicals that may counter cellular disorders \citep{Barrett2006}, including candidates to treat Alzheimer’s, and small cell lung cancer \citep{lamb2006connectivity, sirota2011discovery}. More sophisticated approaches use network analysis and machine learning to efficiently combine drug and disease data \citep{cheng2012prediction, gottlieb2011predict, hoehndorf2013mouse, wu2013computational, bisgin2014phenome}. PREDICT \citep{gottlieb2011predict} is one of most frequently cited computational drug repurposing methods based on machine learning and provides a ranking drug disease associations on the basis of their similarity to a set of known associations. PREDICT has been the subject of 400+ citations (Google Scholar) since it was published in 2011. It has reported a high AUC (0.90) for predicting drug indications, however neither the original data nor the software to produce the results are available to directly replicate this computational study.

\vspace{\baselineskip}

The ability to reproduce original research results are contingent on their availability to the research community. The FAIR principles \citep{wilkinson2016fair}, published in 2016 and now widely endorsed and adopted by funding agencies and publishers, articulate the need for high quality, machine readable data and metadata. Highly diverse communities, from the biomedical sciences to the social sciences and humanities, are now working towards defining their expectations around the publication and sharing of their data. In anticipation, new methods and infrastructure are needed to facilitate the generation of FAIR data. 

\vspace{\baselineskip}

Here, we describe a methodology to publish scientific workflows as FAIR data, whether they involve manual or computational steps, and evaluate it in the context of an open source implementation of the PREDICT drug repositioning methodology. Based on this example in drug repositioning, we will try to answer our research question of how we can use existing vocabularies and techniques to make scientific workflows more open and more FAIR. The main contributions of this paper therefore are (a) general guidelines to make scientific workflows open and FAIR, (b) the OpenPREDICT use case, demonstrating the open and FAIR version of the PREDICT workflow, (c) new competency questions for previously unaddressed reproducibility requirements, and (d) evaluation results on the practicality and usefulness of our approach.

\section*{Background}
The reproducibility crisis is an ongoing methodological issue in which many scientific studies, especially in social and life sciences, are difficult to replicate, a problem that is an important body of research in metascience, particularly for Open Science. For example, in data science, research has shown that data scientists spend 19\% of their time finding, understanding and accessing datasets, and they spend 60\% of their time cleaning and organizing these datasets to use in their studies \citep{CrowdFlower2016}. Therefore, a data scientist spends almost 80\% of the time working on activities that are not the core ones, such as mining data, refining algorithms, building training sets and analyzing the results.  
\subsection*{Reproducibility of scientific workflows}
As data sharing becomes more prevalent, numerous issues must be addressed to ensure the long term success \citep{poldrack2015publication, ioannidis2005most}. Systematic reviews and meta-analyses have recently received increased attention to solve some of these problems \citep{moher2009} and some few initiatives emphasize the methods to share and reuse these workflows. 
The lack of specific details of scientific workflows, including the specification of instructions and their derived code with different steps and parameters, are factors believed to contribute to a non-reproducibility rate of 64\% for pharmacological studies \citep{ioannidis2005contradicted, prinz2011believe}, 89\% in cancer \citep{begley2012drug}, and in 66\% psychology \citep{klein2014investigating}. Usually these workflows are often imprecisely detailed in scientific publications and therefore cannot be reliably reproduced \citep{vasilevsky2013reproducibility}. A recent research over 1.4 million Jupyter notebooks (available in GitHub) found that only 24.11\% of the notebooks could be executed without exceptions and only 4.03\% produced the same results \citep{pimentel2019large}. We also highlight the initiatives under the Common Workflow Language (CWL) \footnote{https://www.commonwl.org/}, which is the de facto standard for syntactic interoperability of workflow management systems.
\par
Several recent initiatives targeted the improvement of workflow reproducibility by introducing methodologies and semantic models to avoid the workflow decay phenomenon \citep{hettne2012best}. The semantic models offer capabilities to represent common understanding of all types of data, including datasets and workflow provenance information, which can be classified as prospective, retrospective and evolution. Retrospective provenance refers to the information about executions, including each atomic event that consumes inputs and produces outputs, as well as information about the execution environment \citep{khan2019sharing}. Prospective provenance refers to the specifications (recipes) that describe the workflow steps and their execution order, typically as an abstract representation of these steps , as well as expected input/output artefacts \citep{Cohen-Boulakia2017}. Workflow evolution refers to tracking the versions of workflows and data, enabling tracing workflow version executions and results. 
Multiple initiatives targeted these three characteristics, having the W3C PROV as the most relevant to address provenance requirements. Furthermore, some research particularly targeted the semantic modelling of workflows, such as the Open Provenance Model for Workflows (OPMW) and P-PLAN, which are extensions of PROV, as well as the CWLProv, which is leveraged by the EDAM ontology, especially used in the bioinformatics community. The ProvBook approach targets Jupyter notebooks as workflows, introducing the Reproduce-me ontology, which overlaps with several predicates of the spar FaBiO ontology. Particularly, the W3C community on machine learning is developing the MLschema ontology, which represents machine learning pipelines as workflows. The PWO targets the workflow of scientific publications, while BPMN targets the specification of processes. The DOLCE top-level ontology is also relevant to provide the foundations for semantic modelling, it provides the Workflow class that is grounded in perdurantism. We also highlight the importance of adopting the Dublin Core (DC) and schema.org elements, which are widespread adopted in the semantic web community. Finally, SHACL plays an important role on the formalization of SPARQL queries that should be executed so the workflow can consume the datasets.
\par
These semantic models were reused in our approach to some extent. In particular, we highlight the role of the Workflow4ever project and the core semantic models produced: Research object (ro), Workflow definition (wfdesc), Workflow execution provenance (wfprov) and the Research Object Evolution Ontology (roevo). The lessons learned from the Worflow4ever project \citep{belhajjame2015using} are important for implementing the FAIR principles towards higher workflow sustainability \citep{mons2018data}. Workflow4ever requirements targeted the three types of provenance: (R0) packaging data and information with workflows (retrospective) (R1) Example data inputs should be provided (prospective); (R2) Workflows should be preserved together with provenance traces of their data results (retrospective); (R3) Workflows should be well described and annotatable (prospective); (R4) Changes/Evolutions of workflows, data, and environments should be trackable (workflow versioning). 
As typical in ontology engineering research, the requirements for the development of the semantic model(s) are derived in competency questions. These questions are then used for the functional validation of the semantic model(s) through SPARQL queries that retrieve the data necessary to answer each question. That approach was also adopted by the Reproduce-me research, which introduced some questions that overlap with Worflow4ever. We highlight here some competency questions that are complementary to our approach to some extent: 
 \par
(\textbf{QW1}) Find the creator of the Workflow Research Object (addressing R3); \par

(\textbf{QW2}) Find the workflow used to generate the gene annotation result reported (addressing R2); \par

 (\textbf{QW3}) Find the inputs used to feed the execution of the workflow that generated a given result (addressing R2); \par
 
 (\textbf{QW4}) Find all workflows that have been modified between the two versions (addressing R4); \par 

 (\textbf{QR1}) What are the input and output variables of an experiment (overlaps with QW2 and QW3); \par

 (\textbf{QR3}) Which are the files and materials that were used in a particular step (overlaps with QW3);\par

 (\textbf{QR5}) What is the complete path taken by a scientist for an experiment (overlaps with QW1);\par

 (\textbf{QR7}) Which are the agents directly or indirectly responsible for an experiment (overlaps with QW1); \par

(\textbf{QR8}) Who created/modified this experiment and when (overlaps with QW1); \par

\subsection*{The FAIR ecosystem}
The FAIR principles describe a minimal set of requirements for data management and stewardship. By adhering to the FAIR data principles, the data produced by a solution can be findable, retrievable, possibly shared and reused and, above all, properly preserved \citep{wilkinson2016fair}. The FAIR data principles are organized in four categories, where each one represents a letter and contains a set of requirements based on best practices from data management and stewardship, applied to both data and metadata. Furthermore, measuring how a digital resource adheres to the FAIR data principles, i.e., the FAIRness level of the resource, is not trivial \citep{wilkinson2017design}. The GO FAIR metrics initiative  aims at defining a set of basic metrics to measure each principle, as well as an approach for the definition of new metrics according to the specificities of the application domain. Therefore, the FAIR metrics should be executed to measure the level of FAIRness of a digital object. 
The main components of the FAIR data ecosystem are data policies, data management plans, identifier mechanisms, standards and data repositories \citep{hodson2018turning}. GO FAIR provides a set of technology specifications and implementations for these components, such as for data repository (FAIR Data Point), and the methodologies for semantic modelling (FAIRification process) and evaluation (FAIR metrics). The FAIR Data Point specification \footnote{https://github.com/DTL-FAIRData/FAIRDataPoint/wiki/FAIR-Data-Point-Specification} guides the use of existing standards and vocabularies, such as OAI-PMH, Dublin Core (DC) and W3C DCAT, for metadata specification for data repositories. Therefore, the FAIR Data Point specification aims to combine existing interoperability standards to meet the FAIR objectives, thus, following the FAIR data principles. The FAIR Data Point specification guides the minimal metadata needed towards dataset findability (F) and it specifies how to declare a data license (A) in order to be reusable (R). 
The FAIRification process specifies the steps required to transform an existing dataset to FAIR data, which leverages on conceptual modelling and RDF implementation. The FAIRification process starts by retrieving the non-FAIR data from the source(s). Then, these datasets should be analyzed to enable the understanding of the data structures and how they are mapped to concepts from the domain. The next step, semantic modelling, is crucial and requires either to create new predicates or to reuse semantic models adherent to the FAIR principles. For example, to represent the dataset about protein-protein interactions in human interactome (a CSV file), Bioportal provides the “protein-protein interaction” class. Then, once the dataset is represented with semantic elements, the dataset is transformed to RDF to make data linkable. The next step is to assign the correspondent license to use the dataset, which may include different licensing types for different elements, e.g. an open license for gene identificator and a restricted license to patient name. Then, metadata definition (e.g., for indexed search) is performed by reflecting the used predicates, i.e., classes and properties from the semantic model(s). The following step is to store the FAIRified data into a FAIR data resource, such as a FAIR Data Point implemented in a triplestore (e.g., Virtuoso and GraphDB). Then, if necessary, the data is combined with other FAIR data, allowing the query of combined data. 
The FAIR principles have received significant attention, but little is known about how scientific protocols and workflows can be aligned with these principles. Making a workflow adherent to the FAIR data principles allows general-purpose open technologies to use machines to act on the data. Particularly, the application of FAIR in healthcare showed that these principles can provide a boost to big data-type applications that require the integration of data coming from different sources, achieving “interoperability without the need to all speak exactly the same language” \citep{Imming2018fair}.
While there is a wide agreement on the FAIR principles, there is currently a lack of methods to integrate them into the daily workflows of scientists.  Some ongoing initiatives are targeting the concept of FAIR Software \footnote{https://github.com/LibraryCarpentry/Top-10-FAIR}, such as the Top 10 FAIR Data \& Software Global Sprint event and the initiative under the Software Sustainability Institute \citep{Neil2018}, which provides a checklist of 10 general actions (best practices) to make a software adherent to the FAIR principles.  For example, it guides the re-use well-maintained libraries / packages rather than reimplementing; and use open, community-standard file formats to address the interoperability principles of FAIR. These initiatives highlight that the FAIR aspects address static data but not the dynamic processes that create and consume such data, and as such they are insufficient to allow researchers to fully address the challenges of data sharing and reproducibility. This is important because the details of a scientific workflow with the different steps and parameters are often imprecisely detailed in scientific publications and therefore cannot be reliably reproduced. Preliminary research to reproduce such studies, e.g., the PREDICT workflow, has revealed disturbing issues in obtaining original data, software, and an inability to reproduce the reported results.

\section*{The FAIR workflows approach}
\label{sec:approach}

In this section we describe the workflow representation requirements, giving focus on the novel parts of our overarching model, such as covering manual steps, different workflow abstraction levels (e.g., protocols as SOPs) and versioning on all these levels. Insufficient focus was given to these needs so far. We describe these requirements as competency questions and the configuration of existing elements from semantic models that is used as an RDF profile to cover these requirements. 

\subsection*{Requirements: competency questions}
We conducted structured interviews with data scientists to formulate relevant user requirements for the reproducibility of workflows (the details of the interviews were given in Appendix 1). Those user requirements were essential to infer the design and implementation of an ontology, particularly as competency questions that it should be able to answer. The main goal of these short interviews was to align some characteristics required for the FAIR workbench (that rely on our ontology) with the needs of the researchers, especially from the drug discovery and repurposing domains.\par

The interviewees stated that they experience many challenges in reproducing their or others' work, such as the lack of details of workflow steps, data cleaning and filtering. In addition to this, the information such as parameters, design details used in reproducing the results could be missing. Furthermore, the definition of manual processes or workflows are usually missing or incomplete. Often, software libraries, packages and versions of tools used are not explicitly recorded. Among the suggestions collected to address these challenges, the data scientists highlighted the need of defining a generic way of computational and manual workflows, the need of making metadata of the datasets accessible, adding richer prospective and retrospective provenance, and allow fine-grained workflow versioning linked to outputs produced during distinct executions. A unanimous recommendation is to allow the separation of the workflow steps from the workflow hyper-parameters, so that one can run the same workflow many times without changing the workflow itself.\par

Some of these recommendations were already addressed by the surveyed semantic models, such as the representation of software environment details (e.g., libraries and packages) in Workflow4ever, CWLProv and Reproduce-me. Therefore, we revised all capabilities of the existing approaches that could be used to address the needs collected from the end user requirements. Furthermore, we searched for the gaps in these semantic models and checked whether critical issues could impede their reuse (e.g., ontology consistency checks). From this study, we conclude that none of the related work could completely address all the requirements together. Here we list a new set of competency questions that our approach targeted:\par

\textbf{CQ1 - Questions about manual steps.}\par

CQ1.1: Which steps are meant to be executed manually and which to be executed computationally? \par

CQ1.2: For the manual steps, who are the agents responsible to execute them (individuals and roles)? \par

CQ1.3: Which datasets were manually handled and their respective formats? \par

CQ1.4: What are the types of manual steps involved? \par

\textbf{CQ2 - Questions about instantiation of general workflows by more specific ones.}\par

CQ2.1: What are the main steps of a general workflow? \par 

CQ2.2: What are the steps of a specific workflow and how are they described? \par

CQ2.3: What higher-level description instantiated a certain workflow step? \par

CQ2.4: Who or what method made the instantiation of a semantic/meta level description of a step into an executable workflow step?\par

\textbf{CQ3 - Questions about versioning of workflows and their executions} \par

CQ3.1: What are the existing versions of a workflow and what are their provenance?  \par

CQ3.2: Which instructions were removed/changed/added from one version to another?  \par

CQ3.3: Which steps were automatized from one version to another? \par

CQ3.4: Which datasets were removed/changed/added for the different versions? \par

CQ3.5: Which workflow version was used in each execution and what was generated? \par

To the best of our knowledge, none of the previous research on semantic modelling of workflows (or protocols/processes) address all these requirements together, in few cases some semantic models only partially cover some questions, as explained in the prior section. \par

\subsection*{Semantic model for FAIR workflows}
From the study of the diverse existing semantic models and other data representations, we designed a conceptual model that explicits our ontological commitments regarding the main elements required to respond the competency questions. In this work we adopted the ontology-driven conceptual modelling approach, which is based on the Unified Foundational Ontology (UFO) and its ontological language OntoUML. In summary, this approach guides the correct classification of a concept by observing its two main aspects: (1) its identity principle, i.e., whether all instances of this concept share the same identity principle: if yes, it is a sortal and non-sortal otherwise; (2) its rigidity principle, i.e., whether an instance of the concept should always instantiate this concept (e.g., a person) or the object can (contingently or not) instantiate a concept (e.g., a student). \par

A \textit{Workflow }is a collective of \textit{Instructions} since its parts have the same functional role in the whole, i.e. each \textit{Instruction} has the same descriptive functional role for the whole (\textit{Workflow}). According to the Merriam-Webster dictionary, the \textit{Instruction }term is commonly used in the plural as an $``$outline or manual of technical procedure$"$  (directions) or $``$a code that tells a computer to perform a particular operation$"$ . Here we do not make an explicit differentiation to other terms that have similar (sometimes the same) definitions, such as process, protocol, procedure, plan specification and standard operating procedure (SOP). We adopted a similar conceptualization as most of the existing semantic models: a \textit{Workflow} is a process description (a plan), i.e., a set of step-by-step instructions where each \textit{Instruction }describes an action intended by agent(s) to achieve goal(s). \par
The aggregation relationship between \textit{Workflow }and \textit{Instruction }is weak because it relies on the ordering of the steps’ execution, which can be sequential or parallel and can repeat instructions. Therefore, we included the notion of \textit{Step}, which is able to be ordered through a \textit{precedes} self-relation and referenced as the first step of a Workflow. A \textit{Step} is described by one \textit{Instruction}, i.e., the \textit{Step} is a pointer to the \textit{Instruction} that describes the action to be performed. Different \textit{Steps} can point to the same \textit{Instruction}, which decouples the \textit{Instruction }and \textit{Step }elements, enabling \textit{Workflows }to reuse \textit{Instructions}. An \textit{Instruction }is written by a particular \textit{Agent} (playing the role of \textit{Writer}) in a certain \textit{Language}, such as a natural language in a SOP (e.g., English) or a programming language in an executable pipeline (e.g., Python). \par
Particularly, we classify a \textit{Step} either as \textit{Manual} or \textit{Computational}, representing how the \textit{Step} should be executed in the \textit{Workflow}. Although this type of definition could also be adopted for representing the \textit{Instruction} element, we decided that the manual or computational capability of an \textit{Instruction} should be derived from its \textit{Language} as an inference. This is due to the fact that a $``$computational$"$  \textit{Instruction}, e.g., an \textit{Instruction }described with Python language, can be used by a \textit{Computational Step}, i.e., enacted in a \textit{Software Development Environment} (compiled or interpreted) to be executed by an \textit{Execution Environment}; or can be used by a \textit{Manual Step}, i.e. requires a human interference to be executed. It is often the case that a data scientist needs to manually execute computational instructions, a capability that some workflow systems have, such as Jupyter Notebook (execution of instructions cell by cell). \par
Similar to P-PLAN and the CWL approach, a \textit{Step }has input and output \textit{Variable(s)}, and the output of \textit{Step SA }can be used as the input of \textit{Step SB }if \textit{SA precedes SB }or \textit{SA }precedes any predecessor of \textit{SB}. Similar to PROV (\textit{Usage }element), we also adopted a way to represent the bindings necessary to describe the use of a \textit{Variable}, aiming at detailing the particular role(s) of each entity participating in the binding. The \textit{Usage }element allows the link between an \textit{Instruction}, a \textit{Variable }and other entity(ies) that are relevant for the binding. For example, the \textit{$``$Download Drugbank dataset$"$  (Step) }produces the output \textit{$``$Drugbank dataset online$"$  (Variable)}, which is a pointer to a particular online distribution of a dataset. The binding \textit{Usage $``$Bind Variable to dataset$"$  relates }to the particular \textit{$``$Drugbank online dataset$"$ } (a \textit{Distribution}), described by the \textit{$``$Download Drugbank dataset$"$ } (\textit{Instruction}). \par
Similar to P-PLAN, Workflow4ever and CWLProv, we adopted an element to represent the execution(s) of each \textit{Step} of a \textit{Workflow}, the \textit{Step Execution}, which \textit{corresponds to} a \textit{Step}. A \textit{Step Execution} can generate \textit{Execution Artifact(s)}, such as a linear regression \textit{Model} that was generated by a machine learning \textit{Workflow}. \par

For the implementation (coding) of the ontology, our approach is leveraged by the best practices on reusing parts of semantic models to build a profile ontology. According to the W3C Profiles Vocabulary\footnote{ https://www.w3.org/TR/dx-prof/ }, a profile is $``$a named set of constraints on one or more identified base specifications or other profiles, including the identification of any implementing subclasses of data types, semantic interpretations, vocabularies, options and parameters of those base specifications necessary to accomplish a particular function". One of the main advantages of this approach is the interoperability enhancement for workflow modelling, since we adopted here the elements of the most appropriate semantic models for workflow representation. Moreover, by adopting the elements of the workflow models, we also (indirectly) address their competency questions. \par
\begin{figure}[ht]
\centering
\includegraphics[width=\linewidth]{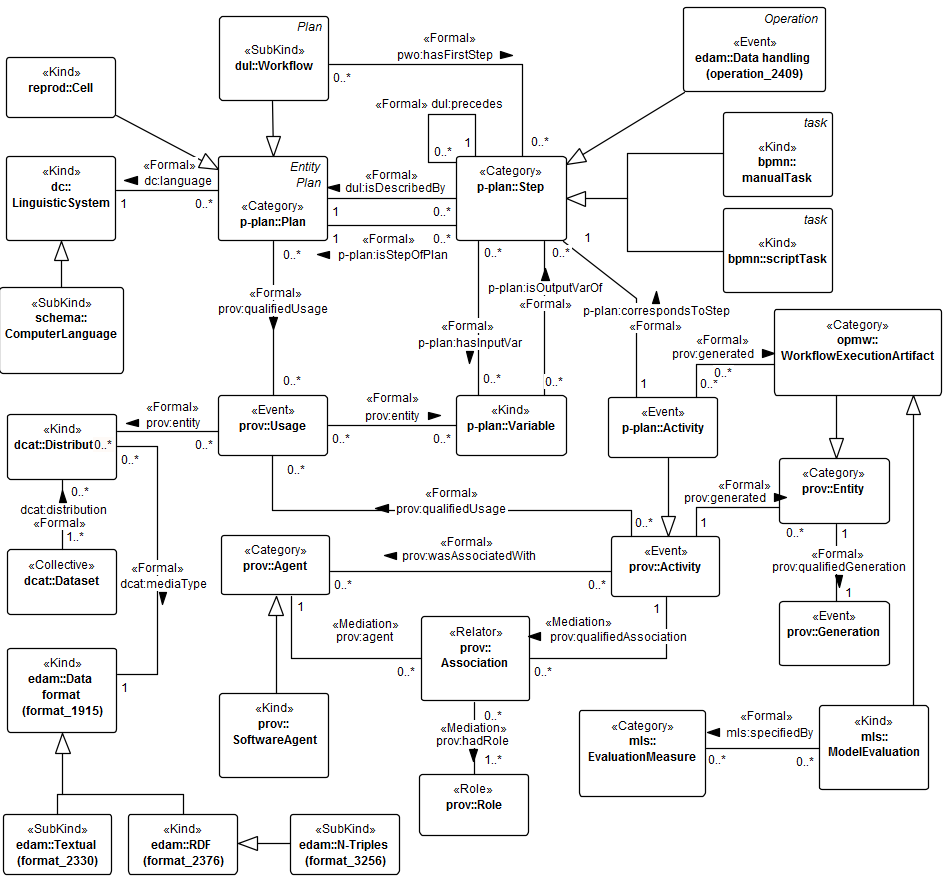}
\caption{Profile ontology for semantic modelling of workflows}
\label{fig:ontology_profile}
\end{figure}

 Figure ~\ref{fig:ontology_profile} illustrates the main elements of the profile\footnote{ https://w3id.org/fair/plex  }, which is grounded mainly on DOLCE Ultra Lite (DUL), W3C PROV, P-PLAN and BPMN 2.0 semantic models. The elements chosen cover each element from our conceptual modelling and target the new competency questions. The most relevant ontology used is P-PLAN because it provides the most adequate abstract terminology to describe plans, i.e., the main building blocks required, besides the wide adoption by several initiatives. Furthermore, P-PLAN is extended from the PROV ontology, which is well-grounded in a high-quality top-level ontology (DOLCE) that facilitates the translations from the conceptual model. Therefore, the DOLCE Ultra-Lite (DUL) ontology also plays a relevant role and provide one of the main elements (\textit{dul:Workflow}). A \textit{dul:Workflow} is a \textit{p-plan:Plan }that represents the top-level of a workflow (or protocol) version, which must be referenced as a whole, e.g., the OpenPREDICT workflow v0.2. Therefore, the intention of \textit{dul:Workflow} is to classify whether the plan is a whole workflow. \par
 
 The \textit{p-plan:Plan} category is the core element of our ontology profile and is the class used to classify any type of instruction. Since the definition of the granularity of an instruction is quite challenging, the \textit{p-plan:Plan} element allows the composition of instruction by means of ordered steps. Therefore, we reused \textit{p-plan:Plan} as a specific type of \textit{prov:Plan} that can be composed by smaller steps (\textit{p-plan:Step}) that have variables (\textit{p-plan:Variable}) as input(s) and/or output(s). The main incorporated classes from P-PLAN are \textit{p-plan:Plan}, \textit{p-plan:Step} and \textit{p-plan:Variable}, while the main incorporated properties are \textit{p-plan:hasInputVar}, \textit{p-plan:hasOutputVar }and \textit{p-plan:isStepOfPlan}. To support the representation of mereology and ordering of steps within a plan, we adopted the \textit{pwo:hasFirstStep} property to indicate the first step of a plan and the \textit{dul:precedes} property to indicate whenever a step precedes another. This choice enables the representation of sequential and parallel steps. \par
 We decouple a particular step within a workflow from its instruction by creating the pattern \textit{p-plan:Step dul:isDescribedBy p-plan:Plan}, where each step always point to one instruction. This approach allows the separation of the workflow steps from the hyper-parameters, enabling the reuse of instructions by different workflows. Therefore, in our approach a step is a lightweight object (like a pointer) that serves only for ordering of instructions without coupling them to the specific workflow. Besides that, we also adopted the \textit{dul:isDescribedBy }property as a self-relationship of \textit{p-plan:Plan}, meaning that an instruction describes another instruction in a different abstraction level. By using this approach the modeller can represent the common software engineering phases of specification and implementation. While the instructions written in a computer programming language mean the implementation of a software, the instructions written in natural language (or pseudo-code) can be used as the specification of the implementation. For example, the first step of the specification of the machine learning pipeline of OpenPREDICT guides the developer to $``$load all features and gold standard$"$  (a \textit{p-plan:Plan}). The implementation of this instruction is done through four instructions (Jupyter notebook cells) ordered via steps. So, these four instructions have together a link (\textit{dul:isDescribedBy}) to the specification as n:1 relationship.\par
 The \textit{prov:Usage} was adopted for the representation of the binding of a variable with other resources, linked from the instruction (\textit{p-plan:Plan}) through the \textit{prov:qualifiedUsage}. Therefore, we adopted this same capability of PROV, enabling the connection to variable(s) (\textit{p-plan:Variable}) and dataset distributions (\textit{dcat:Distribution}). For example, the instruction (\textit{p-plan:Plan}) to $``$download a dataset and save it in the local environment$"$  has a link (\textit{prov:qualifiedUsage}) to the $``$binding the online dataset to a local variable$"$  (\textit{prov:Usage}), which represents the connection between the dataset distribution (\textit{dcat:Distribution}) and the local variable (\textit{p-plan:Variable}) through instances of the \textit{prov:entity} properties. \par
  We adopted the very same profile defined by the FAIR Data Point specification for the representation of datasets (input and output) through the \textit{dcat:Dataset} element, which should be linked to the available distributions (\textit{dcat:Distribution}) through the \textit{dcat:distribution} property, and the URL to download the distribution is represented with \textit{dcat:downloadURL}. We improved this approach with data formats from the EDAM ontology (\textit{edam:format\_1915}) through the \textit{dcat:mediaType} property. A dataset distribution may be available as a text file (\textit{edam:format\_2330}), e.g., a CSV file, and as RDF language (\textit{edam:format\_2376}), e.g., as N-triples serialization (\textit{edam:format\_3256}). For example, the datasets used in our experiments (OpenPREDICT) from DrugBank, Kegg and Sider are provided in N-triples format, while the phenotype annotations dataset is a CSV file.\par
  Our approach also allows the representation of queries over RDF input datasets as SPARQL statements by adopting the SHACL ontology. A \textit{sh:NodeShape} instance represents the constraints that need to be met with respect to a SPARQL query, which is represented with the \textit{sh:SPARQLConstraint} element and linked through the \textit{sh:sparql} property. The use of the SPARQL query over the specific dataset is represented through the \textit{prov:Usage} approach, where a \textit{sh:NodeShape} instance links to a \textit{prov:Usage} with the \textit{sh:targetClass} property. For example, the constraints used to fetch the drug smiles data from the DrugBank are represented with a \textit{sh:NodeShape}, which is linked to the specific SPARQL query (\textit{sh:SPARQLConstraint}) through the \textit{sh:sparql} property, and linked to query execution binding (\textit{prov:Usage}) through the \textit{sh:targetClass}. We adopted the \textit{fabio:Triplestore} element to represent the triplestore where a RDF-based dataset distribution is stored.\par
  We adopted the BPMN 2.0 ontology for the representation of manual and computational activities through the \textit{bpmn:ManualTask} and \textit{bpmn:ScriptTask}, respectively. Therefore, in our ontology we specialized a p-plan:Step as either \textit{bpmn:ManualTask} or \textit{bpmn:ScriptTask} through a generalization set relationship that is disjoint and complete. By adopting this approach, the modeller can represent how instructions (\textit{p-plan:Plan}) should be executed in the workflow. Notice that the same instruction can be executed as a manual and a computational step. For example, a Python instruction often used in Jupyter notebooks is the \textit{variable\_name.head()} method, which prints an excerpt of the variable content (as a data table) so the data scientist can look for any issues with the data before continuing the workflow execution. Although it is an instruction written in a programming language, its execution should be manually, i.e., it requires human intervention to run it. This approach also allows the end user to search for manual steps in a workflow that may be converted to computational steps in a next version. We adopted an extensible approach for the classification of instructions according to existing semantic models that provide elements to describe the instruction(s) of particular workflow systems. For example, we adopted \textit{reprod:Cell }(as a \textit{p-plan:Plan}), which is the element used by ProvBook to represent a cell of a workflow implemented with Jupyter Notebook. \par
  For the representation of retrospective provenance, i.e., information about prior executions of the workflow, we adopted the very same approach of P-PLAN leveraged by PROV. Therefore, the \textit{p-plan:Activity} element, which is a specialization of \textit{prov:Activity}, is chosen for the representation of the execution of each step in a workflow. This is enabled by the use of \textit{p-plan:correspondsToStep }property between a \textit{p-plan:Activity} and a \textit{p-plan:Step}. \par
  The agent associations’ pattern from PROV is used to represent that an association (\textit{prov:Association}) defines the role (\textit{prov:Role}) of an agent (\textit{prov:Agent}) during the execution of a step (\textit{prov:Activity}). With this pattern, it is possible to represent that specific software (\textit{prov:SoftwareAgent}) was (or were) playing the role of execution environment for a workflow. For example, the Jupyter Notebook (\textit{prov:SoftwareAgent}) was used as execution environment (\textit{prov:Role}) for all computational steps of the OpenPREDICT workflow. Furthermore, as a practical design decision, we extended the notion of \textit{prov:Association }for endurants, so the modeller can apply the association pattern similarly to the perdurant way, i.e., use the property \textit{prov:hadPlan} from \textit{p-plan:Association} to \textit{p-plan:Plan} instead of the relation from \textit{prov:Activity} through \textit{prov:qualifiedAssociation}. Therefore, this approach allows the modeller to represent the association of agent roles to an instruction. For example, Remzi is the OpenPREDICT main developer, so the $``$Remzi as developer of OpenPREDICT$"$  (\textit{prov:Association}) links to (a) the $``$Developer$"$  (\textit{prov:Role}) through \textit{prov:hadRole} property, (b) the Remzi object (a \textit{prov:Agent}) through \textit{prov:agent}; and (c) all OpePREDICT instructions (\textit{p-plan:Plan}), through \textit{prov:hadPlan}. Notice that, although the terminology of these properties targeted the perdurant aspect (\textit{prov:Activity}), these properties are also useful for the endurant aspect. Ideally, they should have the adequate endurant terminology, so instead of prov:hadPlan, it should be $``$\textit{prov:hasPlan}$"$  (similarly for \textit{prov:hadRole }too).\par
  One of the main characteristics of the workflow execution approaches is to specify the outputs produced by each execution. We specialized the PROV approach, which uses the prov:generated property from a \textit{prov:Activity} to a \textit{prov:Entity}, by incorporating the \textit{opmw:WorkflowExecutionArtifcat} (as a \textit{prov:Entity}), allowing the use of \textit{prov:generated} from \textit{p-plan:Activity}. Therefore, each step execution can generate workflow execution artifact(s). We also adopted the generation pattern from PROV (\textit{prov:Entity prov:qualifiedGeneration prov:Generation}), which allows the representation of specific aspects of the event, such as its temporal characteristics. our ontology incorporates the MLS ontology for the representation of machine learning workflow specificities, such as the model trained and its evaluation measures. A \textit{mls:ModelEvaluation} is represented as a \textit{opmw:WorkflowExecutionArtifact} once its aim is connecting an evaluation measure specification (\textit{mls:EvaluationMeasure}) to its value. For example, it can be used to specify the accuracy of the models trained during different executions. \par
  For the representation of versioning we adopted a simple approach of applying \textit{dc:hasVersion }property to workflows (\textit{dul:Workflow}), instructions (\textit{p-plan:Plan}), languages (\textit{dc:LinguisticSystem}), software (\textit{prov:SoftwareAgent}), besides the elements guided by the FAIR Data Point specification (e.g., from DCAT ontology). Furthermore, a fundamental characteristic for versioning is the link of the different versions of these elements through the \textit{prov:wasRevisionOf property}.\par

\section*{OpenPREDICT Use Case} 
\begin{figure}[ht]
\centering
\includegraphics[width=\linewidth]{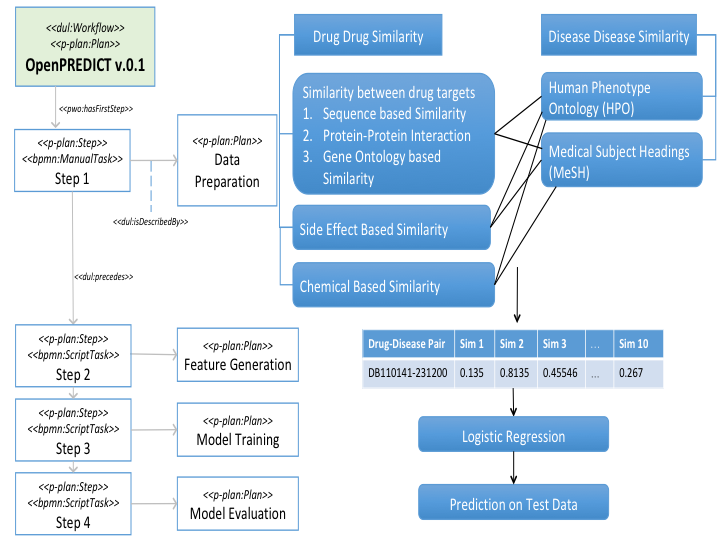}
\caption{ OpenPREDICT Workflow (version 0.1) with manual and computational steps .}
\label{fig:openpredict}
\end{figure}

OpenPREDICT workflow implemented the same steps of the original PREDICT, i.e., 5 drug-drug similarity and 2 disease-disease similarity measures were used to train a logistic classifier to predict potential drug-disease association (see Figure~\ref{fig:openpredict}).  The workflow consists of following main steps:
\begin{enumerate}
    \item \textbf{Data preparation:} In this step, necessary data set is collected and preprocessed. 
     \item \textbf{Feature Generation:} In this step, we generate features from the collected data sets. Drug-drug and disease-disease similarity scores were combined by computing the weighted geometric mean. Thus, we combine 5 drug-drug similarity measures and 2 disease-disease similarity measures, resulting in 10 features.
      \item \textbf{Model Training:} In this step, the generated features from previous step are to be used to train in a simple logistic classifier.
       \item \textbf{Model Evaluation:} This step uses 2 different cross-validation approaches: One where 10 \% of drugs is hidden and one where 10 \% of associations is hidden for test. ROC AUC, AUPR, accuracy, precision and F-score of the classifier on test data are reported.
\end{enumerate}
 
The detailed explanation of feature generation (2) and model evaluation (4) is presented in Appendix 2. The implementation and the workflow description of OpenPREDICT are available at GitHub \footnote{https://github.com/fair-workflows/openpredict}.

\subsubsection*{FAIRified Data Collection}
The first step to make PREDICT open and FAIR was to achieve FAIR input data, which included finding or generating  Linked Data \citep{bizer2009linked} representations. For that, we made use of Bio2RDF \citep{callahan2013ontology}, which is a key Linked Data resource for the biomedical sciences, providing a network of data from several major biological databases.  We have collected data from Linked Data repositories such as Bio2RDF project whenever possible. The drugs, drug targets, drug chemical structure, and drug target sequence were obtained from DrugBank. In addition, The KEGG dataset was used to obtain more drug targets. The SIDER dataset was used for drug side effects and the HGNC dataset was used for gene identifier mapping and gene ontology (GO) annotation. Linked data versions of these datasets (Drugbank, KEGG, HGNC and SIDER) can be downloaded from the Bio2RDF project. We used the supplementary file provided by \citep{menche2015uncovering} for protein-protein interactions and  disease phenotype annotations that link HPO terms to OMIM diseases can be downloaded from \footnote{https://hpo.jax.org/app/download/annotation} . MeSH annotations were collected from \footnote{https://paccanarolab.org/disease\_similarity} \citep{caniza2015network} and annotations were also obtained by NCBO annotator API  \citep{  noy2009bioportal}  using the OMIM disease description.
\par

The data that are not in linked data format (RDF) were converted to RDF with a FAIRification process \cite{jacobsen2019generic}. We stored the collected datasets in a triplestore and created SPARQL queries to access the triplestore in order to produce the features for PREDICT's method.

\par 

The OpenPREDICT workflow has two versions (0.1 and 0.2). In the first, we experimented the FAIRifier tool with the two inputs that are provided as text files (CSV), i.e., \textit{human-interactome-barabasi.tab} (protein-protein interactions in human interactome) and \textit{phenotype\_annotation\_hpoteam.tab} (disease phenotypic descriptions). Besides the formalization of the manual steps through our approach, we also provide guidelines for the manual steps. In the second, we wrote Python scripts for FAIRificiation process of these datasets, evolving most of the manual steps to computational ones. \par 
The process starts with obtaining the raw data from the original sources and making a local copy of raw data. The next step is to clean data if necessary and define a semantic model for the data. The semantic model should ideally be defined with existing ontologies that can be found in ontology repositories. In order to find appropriate ontology and terms, the ontology repositories such as BioPortal for biomedical data can be used for search. If no suitable ontology was found, a vocabulary with its description should be created. We next define metadata that includes the provenance and the licence with HCLS dataset description. \par

For example, Bioportal was used to find the $``$protein-protein interaction$"$  class, and the element $``$Protein interactions$"$  from EDAM (\textit{edam:topic\_0128}) was selected to represent the \textit{human-interactome-barabasi.tab} (protein-protein interactions in human interactome) dataset (CSV file). This dataset contains 3 columns: first and second columns provide a gene identifier, each one representing a gene interactor, i.e. the role of the gene in the interaction; and the third column the source. So we searched for the $``$interactor$"$  and $``$source$"$  concepts, choosing the properties \textit{bio2rdf:interactor\_a}, \textit{bio2rdf:interactor\_b }and \textit{bio2rdf:source}, respectively. Similarly, for the disease phenotype descriptions dataset, we also applied the FAIRifier approach. In this case, the dataset provides 14 columns and we selected 3: column 1 representing the source and column 2 representing phenotypes. We represented the dataset with \textit{sio:Disease} (SIO\_010299), column 1 with \textit{bio2rdf:source} and column 2 with \textit{sio:hasPhenotype} (SIO\_001279).\par

Besides the execution of the FAIRification process in the datasets, we also applied the profile ontology to formalize the FAIRification steps. The RDF snippet below shows the representation of the FAIRifier step of the protein-protein interaction dataset in OpenPREDICT v0.1:\par

\begin{lstlisting}[label=lst:lst11, breaklines=true, basicstyle=\small\tt]
opredict:Step_Execute_FAIRifier_process_to_human_interactome_barabasi
  rdf:type bpmn:ManualTask ;
  rdf:type p-plan:Step ;
  p-plan:isStepOfPlan opredict:Plan_Main_Protocol_v01 ;
  dul:isDescribedBy opredict:Plan_Execute_FAIRifier_process_to_human_interactome_barabasi ;
  rdfs:label "Execute FAIRifier process to human interactome barabasi" ;
\end{lstlisting}

\subsection*{OpenPREDICT workflow representation}
The diagram (Figure 4.1) illustrates the main steps of the OpenPREDICT workflow, in which the Main Protocol is represented as a \textit{dul:Workflow} and a \textit{p-plan:Plan}, with version set through the \textit{dc:hasVersion} property. Current, OpenPREDICT has two versions (0.1 and 0.2) and, therefore, two instances of the Main Protocol were created. The workflow is represented through 4 \textit{p-plan:Step}: data preparation, feature generation, model training and evaluation, and presentation of results. Each one is defined by (\textit{dul:isDescribedBy}) its homonymous \textit{p-plan:Plan}. In the first version of OpenPREDICT all steps within data preparation were manual (\textit{bpmn:ManualTask}), as the FAIRification process and the preparation steps on data that were already provided as RDF. The second version of OpenPREDICT automated most of these manual steps, requiring less human intervention. \par

Among the main patterns that we observed when representing OpenPREDICT versions with our ontology, we highlight:\par
\vspace{\baselineskip}
\textbf{1- Prospective provenance:} decoupling the workflow step from the instruction, linking the \textit{p-plan:Step} to \textit{p-plan:Variable} through \textit{p-plan:hasInputVar} and \textit{p-plan:hasOutputVar}, while the \textit{p-plan:Plan }links to the \textit{prov:Usage }through \textit{prov:qualifiedUsage }property, describing how to bind the variable to other resources. Example:\par

\begin{lstlisting}[label=lst:lst11, breaklines=true, basicstyle=\small\tt]

opredict:Step_Download_Drugbank_dataset
  rdf:type bpmn:ManualTask ;
  rdf:type edam:operation_2409 ;
  rdf:type p-plan:Step ;
  p-plan:hasOutputVar opredict:Variable_Drugbank_dataset_online ;
  p-plan:isStepOfPlan opredict:Plan_Main_Protocol_v01 ;
  dul:isDescribedBy opredict:Plan_Download_Drugbank_dataset ;
  dul:precedes opredict:Step_Save_Drugbank_dataset ;
  rdfs:label "Download Drugbank dataset" ;
.

opredict:Plan_Download_Drugbank_dataset
  rdf:type p-plan:Plan ;
  dc:description "Download Drugbank dataset" ;
  dc:language :LinguisticSystem_xsd_language_English ;
  rdfs:label "Download Drugbank dataset" ;
  prov:qualifiedUsage opredict:Usage_Fetch_download_Drugbank_dataset_to_variable ;
.

opredict:Usage_Fetch_download_Drugbank_dataset_to_variable
  rdf:type prov:Usage ;
  rdfs:label "Link variable to download Drugbank dataset" ;
  prov:entity opredict:Distribution_release-4-drugbank-drugbank.nq.gz;
  prov:entity opredict:Variable_Drugbank_dataset_online ;
.

opredict:Distribution_release-4-drugbank-drugbank.nq.gz
  rdf:type dcat:Distribution ;
  rdfs:label "release/4/drugbank/drugbank.nq.gz" ;
  dcat:downloadURL "http://download.bio2rdf.org/files/release/4/drugbank/drugbank.nq.gz" ;
  dcat:mediaType opredict:DataFormat_nq_compressed_gz ;
.

opredict:Variable_Drugbank_dataset_online
  rdf:type p-plan:Variable ;
  rdfs:label "Drugbank dataset online" ;
.

\end{lstlisting}
\vspace{\baselineskip}
\textbf{2- Retrospective provenance:} allowing the representation of executions and output generation, where a \textit{p-plan:Activity} is linked to a \textit{p-plan:Step} through the \textit{p-plan:correspondsToStep} property and to the outputs (\textit{opmw:WorkflowExecutionArtifact) }through \textit{prov:generated}. Each output has a value (e.g., accuracy rate) and is linked to \textit{prov:Generation} through \textit{the prov:qualifiedGeneration} property, which at least specifies when the generation occurred (\textit{prov:atTime}). Example: \par

\begin{lstlisting}[label=lst:lst11, breaklines=true, basicstyle=\small\tt]
opredict:Activity_Model_preparation_train_and_evaluation_Execution_1546302862
  rdf:type p-plan:Activity ;
  p-plan:correspondsToStep opredict:Step_Model_preparation_train_and_evaluation ;
  prov:generated opredict:ModelEvaluation_Accuracy_Execution_1546302862 ;
  prov:generated opredict:ModelEvaluation_AveragePrecision_Execution_1546302862 ;
  prov:generated opredict:ModelEvaluation_F1_Execution_1546302862 ;
  prov:generated opredict:ModelEvaluation_Precision_1546302862 ;
  prov:generated opredict:ModelEvaluation_Recall_Execution_1546302862 ;
  prov:generated opredict:ModelEvaluation_RocAuc_Execution_1546302862 ;
.

opredict:ModelEvaluation_Accuracy_Execution_1546302862
  rdf:type mls:ModelEvaluation ;
  dc:description "0.833336" ;
  mls:specifiedBy opredict:EvaluationMeasure_PredictiveAccuracy ;
  prov:qualifiedGeneration opredict:Generation_Execution_1546302862 ;
.

opredict:Generation_Execution_1546302862
  rdf:type prov:Generation ;
  prov:atTime "2019-01-01T00:02:31.011"^^xsd:dateTime ;
\end{lstlisting}
\textbf{3- Versioning of workflows:} allowing to track the modification through the \textit{dc:hasVersion} properties on \textit{dul:Workflow}, \textit{p-plan:Plan}, \textit{dc:LinguisticSystem} and \textit{prov:SoftwareAgent}. Furthermore, a fundamental characteristic for versioning is the link of the different versions of these elements through the \textit{prov:wasRevisionOf property}.  Example: \par

\begin{lstlisting}[label=lst:lst11, breaklines=true, basicstyle=\small\tt]
opredict:Plan_Main_Protocol_v02
  rdf:type p-plan:Plan ;
  rdf:type dul:Workflow ;
  dc:created "2019-05-15" ;
  dc:creator opredict:Agent_Remzi ;
  dc:description "OpenPREDICT Main Protocol v.0.2" ;
  dc:hasVersion "0.2" ;
  dc:language :LinguisticSystem_xsd_language_English ;
  dc:modified "2019-07-03" ;
  pwo:hasFirstStep opredict:Step_Prepare_Input_Data_Files_v02 ;
  rdfs:label "Main Protocol v.0.2" ;
  prov:wasAttributedTo opredict:Agent_Remzi ;
  prov:wasRevisionOf opredict:Plan_Main_Protocol_v01 ;
.

opredict:Plan_Main_Protocol_v01
  rdf:type p-plan:Plan ;
  rdf:type dul:Workflow ;
  dc:created "2018-11-27" ;
  dc:creator opredict:Agent_Remzi ;
  dc:description "OpenPREDICT Main Protocol v.0.1" ;
  dc:hasVersion "0.1" ;
  dc:language :LinguisticSystem_xsd_language_English ;
  dc:modified "2019-05-15" ;
  pwo:hasFirstStep opredict:Step_Prepare_Input_Data_Files ;
  rdfs:label "Main Protocol v.0.1" ;
  prov:wasAttributedTo opredict:Agent_Remzi ;

.
\end{lstlisting}

\section*{Evaluation}
In this section we describe the overall evaluation strategy of our approach in twofold: First, we revisit each FAIR principle and explain how the principle is addressed. Second, we applied the traditional ontology validation methodology by answering the competency questions through the execution of SPARQL queries (available in Appendix 3). \par

\subsection*{Addressing the FAIR principles}
In order for our workflow to comply with FAIR principles, we checked each FAIR criterion defined in \citep{wilkinson2016fair}, as identified between parentheses below. First, global and persistent identifiers were assigned to resources defined in the workflow and associated data. Rich metadata for workflow and input and output data were created using HCLS and FAIR data point specification (F2). In addition, the metadata we generated contain an explicit global and persistent identifier of the data they describe (F3). In order to enable the workflow and the data used to be searched, they were uploaded in a triple-store as a FAIR Data Point.  Data can be queried through SPARQL over HTTP(S) protocol (A1.1). Since the data is not private or protected, we don't require authentication and authorisation mechanism (A1.2). Metadata will be published on another registry to make the metadata accessible even the data is no longer available (A2). We used RDF, OWL with commonly used controlled vocabularies, ontologies  such as bio2rdf vocabulary, SIO and PROV to model input data and workflows (I1). HCLS dataset specification and FAIR Data Point specification were used to define the metadata and provenance of data (I2). Meaningful links between (meta)data such bio2rdf links and data and workflow were created (I3).  We describe the semantic model of the workflow and the data with a commonly used vocabulary such as ML-Schema and Re-Produceme (R1). We provide the license (R1.1) and provenance information in the metadata using FAIR data point specification (R1.2), and HCLS specification (R1.3) and PROV.

\subsection*{Responding competency questions}
Besides evaluating whether each FAIR principle was addressed, we also assessed the profile ontology using the common semantic validation approach, which is based on SPARQL queries used to respond the competency questions. All questions listed in Section 3 could be answered by running the SPARQL queries over the OpenPREDICT use case. The complete queries and results obtained from OpenPREDICT triplestore can be found in Appendix 3. Therefore, the reproduction of this validation can be performed by re-executing the queries on the RDF representation of the OpenPREDICT workflow. Below we explain the result for each competency question. 

\subsubsection*{CQ1 - Questions about manual steps.}

\paragraph*{CQ1.1: Which steps are meant to be executed manually and which to be executed computationally? }
The SPARQL query we built to answer this question (see appendix) first filters all steps within the first version of OpenPREDICT workflow (\textit{opredict:Plan\_Main\_Protocol\_v01}). The results show each step and its type - manual (\textit{bpmn:ManualTask}) or computational (\textit{bpmn:ScriptTask}) - as well as the respective instructions (\textit{p-plan:Plan})\ that describe the steps. In summary, OpenPREDICT v0.1 has 28 manual steps and 14 computational steps (42 in total), while v0.2 has 9 manual steps and 9 computational steps (18 in total). This difference reflects the automatization of most of the manual steps within data preparation (evolving from manual to computational) and the simplification of the computational steps described in fewer Jupyter Notebook cells.  \par


\paragraph*{CQ1.2:For the manual steps, who are the agents responsible to execute them? }
To answer this question we filtered the results for only manual steps through the statement:\par
\textit{values ?stepType $ \{ $  bpmn:ManualTask $ \} $ }\par

The result is a list of all steps and roles related to each one, such as executor, creator, developer, and publisher. For example, Remzi is creator, developer and executor of all instructions, while Ahmed is developer of some computational steps and Joao is the executor of the entire OpenPREDICT workflow. This approach allows the representation of multiple roles played by different agents within each step.\par

Similar to the Reproduce-me approach (query QR7), our ontology leverages on the PROV ontology to address the different types of agents and roles through the\textit{ prov:wasAttributedTo} property. As in Workflow4ever and Reproduce-me approaches (queries QW1 and QR5), we also apply the \textit{dc:creator} and \textit{dc:publisher }properties for the direct relation from an instruction to an agent. Therefore, our ontology enables the representation of common roles through these properties, but also allows the representation of multiple roles (e.g., developer, executor, researcher) by adopting the \textit{prov:Association} pattern. Additionally, our ontology leverages on the Reproduce-me (query QR8) approach for creation and modification dates (\textit{dc:created }and\textit{ dc:modified }properties). \par


\paragraph*{CQ1.3: Which datasets were manually handled and what are their formats? }
OpenPREDICT v0.1 computational steps use 5 datasets, as explained in Section 4, that required manual steps to be prepared: Phenotype annotations, human interactome barabasi, Drugbank, Kegg and SIDER. OpenPREDICT v0.2 computational steps use more 2 datasets: the gold standard drug indications and the mesh annotations (disease similarity), which data preparation steps were also automated. Table~\ref{table:FAIRifiedDataset} summarizes the result with the list of all datasets used in version v0.2. The main elements of the query reflect the FAIR data point specification with DCAT elements (\textit{dcat:Distribution, dcat:downloadURL} and \textit{dcat:mediaType}), PROV (\textit{prov:Usage }and \textit{prov:qualifiedUsage}) and EDAM classification for data handling steps (\textit{edam:operation\_2409}) and data formats (media types). 

\begin{table}[]
\resizebox{\textwidth}{!}{%
\begin{tabularx}{\textwidth}{|X|X|X|X|}
\hline
\textbf{Dataset file (dcat:Distribution)}                                  & \textbf{Data format}        & \textbf{Download URL}                                                                                                \\ \hline
opredict:Distribution \_gold\_standard\_drug\_indications \_msb201126-s4.xls & .tab with tabular separator & https://www.ncbi.nlm.nih.gov /pmc/articles/PMC3159979 /bin/msb201126-s4.xls                                            \\ \hline
                                                                         
opredict:Distribution \_mesh\_annotation\_mim2mesh.tsv                      & .tab with tabular separator & http://www.paccanarolab.org/ static\_content/disease \_similarity/mim2mesh.tsv                                         \\ \hline

opredict:Distribution\_phenotype \_annotation\_hpoteam.tab \_Build\_1266     & .tab with tabular separator & http://compbio.charite.de/ jenkins/job/hpo.annotations/1266/ artifact/misc/phenotype\_annotation\_hpoteam.tab          \\ \hline
            
opredict:Distribution\_pubchem \_to\_drugbank\_pubchem.tsv                  & .tab with tabular separator & https://raw.githubusercontent.com/ dhimmel/drugbank/ 3e87872db5fca5ac427 ce27464ab945c0ceb4ec6/ data/mapping/pubchem.tsv \\ \hline
                                                                
opredict:Distribution \_release-4-kegg-kegg-drug.nq.gz                      & .nq (RDF) compressed as .gz & http://download.bio2rdf.org/ files/release/4/kegg/kegg-drug.nq.gz                                                     \\ \hline
               
opredict:Distribution \_release-4-sider-sider-se.nq.gz                      & .nq (RDF) compressed as .gz & http://download.bio2rdf.org/ files/release/4/sider/sider-se.nq.gz                                                     \\ \hline
                
opredict:Distribution \_srep-2016-161017-srep35241-extref-srep35241-s3.txt  & .txt with comma separator   & https://media.nature.com/ full/nature-assets/srep/ 2016/161017/srep35241/ extref/srep35241-s3.txt                       \\ \hline
\end{tabularx}%
}

\caption{All datasets used in OpenPREDICT version v0.2}
\label{table:FAIRifiedDataset}
\end{table}

\paragraph*{CQ1.4: What are the types of manual steps involved, and what are their inputs and outputs? }
Similar to the Reproduce-me approach (query QR1 and QR3), our ontology leverages on the P-PLAN ontology to address the variables used as input and output of the manual steps, mostly during data preparation in OpenPREDICT v0.1, such as downloading and saving the datasets listed in the results of CQ1.3. For example, the input of \textit{opredict:Step\_Save\_files\_in\_triplestore} are variables that indicate the local file of each dataset (serialized as RDF) and the output variable indicating the endpoint to upload all datasets (\textit{opredict:Variable\_Triplestore\_endpoint\_for\_input\_data}). \par

When changing the filter from manual steps to computational steps, the pattern followed was to classify the output variables of a step (a Jupyter Notebook cell) according to the data saved in files. For example, in feature generation, the \textit{opredict:Step\_Feature\_generation\_01\_Pipeline\_Source\_Cell11} has an output variable for drug fingerprint similarity, indicating the generation of the file $``$drugs-fingerprint-sim.csv$"$ .\par

\subsubsection*{CQ2 - Questions about instantiation of general workflows by more specific ones.}
\paragraph*{CQ2.1: What are the main steps of a general workflow? }
OpenPREDICT workflow follows the common machine learning pipeline process of: data preparation, feature generation, model training, model evaluation and presentation of results. The query returns these steps by looking for the first step of the workflow (through \textit{pwo:hasFirstStep}) and following the preceding path in a recursive way, e.g., 
 \begin{lstlisting}[label=lst:lst11, breaklines=true, basicstyle=\small\tt]
?step1 dul:precedes ?step2.
?step2 dul:precedes ?step3.
?step3 dul:precedes ?step4. (until there is no preceding steps)
\end{lstlisting}
 
 The classification of the step is given by the EDAM specializations of the Operation concept \textit{(operation\_0004}), such as \textit{Data Handling} for data preparation (\textit{edam:operation\_2409}). For the sake of simplicity, model training and evaluation were performed within the same step. The main steps are listed below:

  \begin{lstlisting}[label=lst:lst11, breaklines=true, basicstyle=\small\tt]
opredict:Step_Prepare_Input_Data_Files
opredict:Step_Feature_generation_Pipeline_OpenPREDICT_ipynb
opredict:Step_Model_preparation_train_and_evaluation_Workflow_OpenPREDCIT_-_ML_ipynb
opredict:Step_Format_results_for_presentation

\end{lstlisting}

\paragraph*{CQ2.2: What are the steps of a specific workflow? }
Similar to the previous question, the SPARQL query uses the properties that allow the ordering of steps’ execution (\textit{pwo:hasFirstStep} and \textit{dul:precedes}). The pattern \textit{p-plan:Step dul:isDescribedBy p-plan:Plan} allows answering this question, by representing how a step is described by an instruction. This pattern resembles the one used by Workflow4ever (QW2), which applies the \textit{wfdesc:hasWorkflowDefinition} (\textit{dul:isDescribed}) to link a \textit{wfdesc:Workflow} (\textit{p-plan:Step}) to a \textit{wfdesc:WorkflowDefinition }(\textit{p-plan:Plan}), aiming at representing the instructions (e.g., a Python script) that are natively understood by the \textit{wfdesc:WorkflowEngine} (\textit{prov:SoftwareAgent}). However, different from this approach, we classify the instruction language (\textit{p-plan:Plan dc:language dc:LinguisticSystem}), allowing the representation of instructions that follow either computer language and/or natural language, which includes pseudo-code - commonly used to specify algorithms before implementing in a particular computer language.\par
The results show that OpenPREDICT has 78 steps in total, where 60 steps belong to v0.1 and 18 belong to v0.2, each step linked to an instruction. 9 instructions were reused from v0.1 to v0.2 regarding data preparation, thus, v0.2 presents 9 new instructions that are used to automate the data preparation phase. These instructions are written as either English (natural language) or Python 3.5 (computer language), where most of the Python ones refer to the Jupyter notebook cells for feature generation and model training and evaluation. \par
\paragraph*{CQ2.3: What higher-level description does a certain workflow step instantiate?}
The SPARQL query to answer this question includes the pattern \textit{p-plan:Plan dul:isDescribedBy p-plan:Plan}, which extends the capability described in the previous question, i.e. decoupling steps from instructions, enabling the representation of different abstraction levels of instructions and their relations. This pattern resembles the links between specification artefacts (e.g., conceptual model, activity diagrams and use cases) and implementation artefacts (e.g., software code, deployment procedures and automated tests) in software engineering. Usually, a specification artefact aims at describing the instructions necessary to enable a programmer to create the software code, sometimes automatically generated as in model-driven engineering. For example, a pseudo-code within an activity diagram (\textit{p-plan:Plan}) may describe the behaviour expected (\textit{dul:isDescribed}) for the algorithm behind a service endpoint, which may be implemented as a Python script (\textit{p-plan:Plan}). \par
OpenPREDICT did not formally followed the specification phase of software engineering since it is a research project, having the code developed from the data scientist interpretation perspective about publications related to PREDICT. In research-oriented data science this type of approach is common. However, we created some examples of the pattern that represent the specification of OpenPREDICT workflow. Therefore, the results of this query include 10 Jupyter Notebook cell instructions (\textit{p-plan:Plan}), representing implementation artefacts, that were specified (p-plan:isDescribedBy) by 3 specification instructions (\textit{p-plan:Plan}). The level of abstraction can be derived from the properties of the instruction. For example, the 10 Jupyter Notebook cell instructions were written (\textit{dc:language}) in Python 3.5 (\textit{schema:ComputerLanguage}), while the 3 specification instructions were written in English (\textit{en} value of \textit{xsd:language}). Furthermore, this approach enables links of \textit{s }(specification artefacts)\textit{ x i }(implementation artefacts), where \textit{i>s}, i.e., a specification artefact usually describes several software code lines (instructions). In OpenPREDICT, the first specification instruction guides the load of input datasets, which is linked to cells 1-5 of the feature generation step, while the second guides the calculation of scores between pairs of drugs and compute similarity feature, which is linked to cells 6-9. \par

\subsubsection*{CQ3 - Questions about versioning of workflows and their executions}
\paragraph*{CQ3.1: What are the existing versions of a workflow and what are their provenance? }
The collective workflow (the whole) is represented as a \textit{dul:Workflow} and a \textit{p-plan:Plan}. Similar to other approaches (Workflow4ever, Reproduce-me, CWLProv, among others) the query to answer this question makes use of DC properties (e.g., \textit{dc:creator, dc:created, dc:modified}) and PROV (e.g. \textit{prov:wasAttributedTo}) for prospective provenance. It also covers workflow versioning through \textit{dc:hasVersion }and \textit{prov:wasRevisionOf}, where the former is responsible for version of \textit{dul:Workflow} and the latter to link an instruction to another (\textit{p-plan:Plan prov:wasRevisionOf p-plan:Plan }pattern). The retrospective (executions) provenance is supported by the link from an execution (a \textit{p-plan:Activity}) to the correspondent step (\textit{p-plan:correspondsToStep} property), which is a pattern that resembles most of the aforementioned semantic models. The main difference here is the assumption that any instruction (\textit{p-plan:Plan}) should be versionable, thus, all executions link to a versioned instruction. Differently from Workflow4ever approach (QW4), here we do not introduce any elements regarding the specification of the changes (e.g., \textit{roevo:ChangeSpecification}). The results for OpenPREDICT show 2 workflows (v0.1 and v0.2), both created by and attributed to Remzi, where v0.2 links to the prior version (v0.1). \par
\paragraph*{CQ3.2: Which instructions were removed/changed/added from one version to another? }
3 SPARQL queries were written to answer whether the instructions of OpenPREDICT v0.1 were removed or changed or added in v0.2. Each SPARQL uses the identificator of the workflow versions (retrieved in CQ3.1) as an input parameter to perform the comparison from one version to another. For the query for removed instructions, it considers all instructions used in v0.1 that are not used in v0.2 and excludes the instructions that were changed. For the query for changed instructions, it considers the instructions with the \textit{prov:wasRevisionOf }property. For the query for added instructions, the SPARQL query uses the reverse logic from the removed.\par
47 instructions were removed from v0.1 to v0.2 due to the refactoring of the code of feature generation, model training and model evaluation, and the elimination of several manual steps in data preparation. 3 instructions were changed, reflecting the porting of the FAIRification manual steps to computational steps in data preparation, i.e., download and save human interactome barabasi, and phenotype annotations. 7 instructions were added in v0.2, where 3 of them represent the new Python scripts for data preparation of the new data sources, other 3 represent the new scripts for feature generation and the remaining for model training. \par
\paragraph*{CQ3.3: Which steps were automatized from one version to another? }
This query is quite similar to the one used for changed instructions (CQ3.2) but it makes explicit that the old version of the instruction used as manual step (\textit{bpmn:ManualTask}) was modified to an instruction used as computational step (\textit{bpmn:ScriptTask}) in the new version. The results confirm the findings from the previous query regarding the 3 instructions that were ported from manual steps to computational steps, namely the data preparation top-level instruction, the FAIRification instructions (download and save human interactome barabasi, and phenotype annotations). Although our approach covers change management, we face the same challenges regarding the dependency of the developer practices for code versioning. This means that, for example, a developer is free to choose whether to remove files from an old version of the software implementation and add files to the new version, even though these files refer to the same capability or routines. Most of the version controls track the changes when the files (old and new) have the same name and path (i.e., the step identificator), which is a similar approach used here. \par
\paragraph*{CQ3.4: Which datasets were removed, changed, oradded from one version to the next? }
This question can be answered by mixing the same query of CQ1.3 (datasets manually used) with the logic used in query CQ3.2, i.e., one SPARQL query to the datasets removed, one for the changed and one for the added. The query results over OpenPREDICT (v0.1 and v0.2) confirm the findings of CQ1.3, where none datasets were removed from the old version to the new, none changed and 2 were added. \par
\paragraph*{CQ3.5: Which workflow version was used in each execution and what was generated? }
This question is answered by exploiting the pattern \textit{p-plan:Activity p-plan:correspondsToStep p-plan:Step}, where the step is part of the dul:Workflow that provides the workflow version. The OpenPREDICT workflow had 14 executions represented with our profile, exemplifying the execution of some computational steps, i.e., each one a particular Jupyter Notebook cell. Therefore, this approach allows the representation of multiple executions of each step according to the version of the corresponding instruction. Each execution inherits the properties of \textit{p-plan:Activity}, e.g., the event start and end time points. Furthermore, each execution is associated to the correspondent generated artefacts through the \textit{p-plan:Activity prov:generated opmw:WorkflowExecutionArtifact} pattern, a similar approach of Workflow4ever (QW3), which applied the inverse property \textit{prov:wasGeneratedBy}. An artefact generated by an execution can be an evaluation measure of the model trained, such as the model accuracy and recall for that particular execution, i.e., a \textit{mls:ModelEvaluation}. Therefore, OpenPREDICT executions generated the values about the model evaluation measures of accuracy, average precision, F1, precision, recall and ROC AUC. For example, the results show that the model accuracy of v0.1 is 0.83, while v0.2 is 0.85. \par
This query can be further detailed by considering the particular version of each instruction that the executed step implements. In addition, ideally, each output of a Jupyter Notebook cell should be represented as a \textit{opmw:WorkflowExecutionArtifact}, so all generated outputs are stored (similar to ProvBook/Reproduce-me approach). This query can be easily changed to provide aggregations for related analytical questions, such as how often each workflow version was executed.\par

\section*{Discussion}

\subsection*{Reproducibility Challenges}
Our study was unable to fully reproduce the dataset and accuracy of the method reported in the PREDICT paper. The performance results of this study are lower than originally reported. The PREDICT paper reported an AUC of 0.90 in cross-validation, but using the same gold standard, we could only achieved a lower AUC of 0.83. 

We were also able to obtain the drug and disease similarity matrices used in PREDICT from the authors via email request.  Given 5 drug-drug similarity measures for 593 drugs and 2 disease-disease similarity measures for 313 diseases, there are resulting 10 features of combined drug-disease similarities. The logistic classifiers were trained with these pre-computed similarity scores and an average AUC of 0.85 was obtained from 10 repetitions in a 10-fold cross-validation scheme. This is still a significant difference from the AUC of 0.90 what the authors reported in PREDICT study. This indicates  that there was more likely an error in the design or implementation of evaluation, not the aggregation data and calculation of drug-drug and disease-disease similarity scores.

During reproducing the PREDICT study, we faced the following challenges.
\begin{enumerate}
    \item \textbf{Insufficient documentation}
    Many details such as the design of the experiment, data sets, parameters and environment should be noted down so that the same exact results can be reproduced. A scientific article usually is limited by its different organization style and constraints (such as page constraints) of journals and conference proceedings, leading to absence of documented workflow, computational and manual steps. In PREDICT, many details such as the calculation of features were not clearly defined and there was no code provided. 
     \item \textbf{Inaccessible or missing data}
     The data that were made available by the authors could be absent or no longer accessible for many reasons. Without sufficient data, an experiment could not be reproduced. Since no data except the gold standard  data (drug-disease associations) were given, the features for the PREDICT were constructed using openly accessible datasets such as DrugBank and KEGG and SIDER. 
     \item \textbf{Versioning and change of data}
     The results or hypotheses could change as a result of updating the data. In order to draw the same conclusion, it is important to record the version or the date of the data that were collected to draw conclusions in a study. Publicly accessible datasets have increasingly been used to construct models and validate hypotheses for prediction of drug indications. In PREDICT, publicly accessible datasets have been used to construct models and validate hypotheses for prediction of drug indications and drugs were identified by their Drugbank IDs. However, Drugbank IDs are dynamic and subject to change, and the original values may change throughout the time. For example, two drugs (DB00510, DB00313) in the original dataset were merged to the same drug within the current version of the Drugbank.
     \item \textbf{Execution environment and third-party dependencies}
       Often, software libraries, packages and tools version used in a workflow are not explicitly recorded, besides the fact that they might not be maintained or updated with no access to previous releases used in the original workflow. In PREDICT study, version of some libraries library such as the library for semantic similarity calculation were not recorded.
\end{enumerate}

\subsubsection*{Foundations of FAIR Workflows semantic profile}
The reuse of existing semantic models for the representation of our profile showed to be a challenging task. There are several existing semantic approaches to represent workflows that present reproducibility issues and different conceptualizations, sometimes overlapping in their terminology. For example, the prospective part of Workflow4ever implementation (wfdesc\footnote{ https://raw.github.com/wf4ever/ro/0.1/wfdesc.owl }) has consistency issues such as missing disjointness and licensing elements, besides not conforming to the documentation (e.g., for all elements related to workflow templates). On the other hand, semantic models like DUL, W3C PROV and P-PLAN presented higher quality and common foundations (in DOLCE), being easier to reuse and extend. \par
A question that may arise is whether it would be better to create a new ontology rather than creating a profile. We believe that high quality semantic models should be reused, taking benefit from the lessons learned. Furthermore, we consider that reusing existing semantic workflow models really improve semantic interoperability, while creating a new ontology may impede interoperability if it is not accompanied with alignments to the existing semantic models. Therefore, our approach leads to an improved semantic interoperability. Because we reused several semantic models, the competency questions that they target are potentially addressed by our approach. For example, the gap in our approach regarding the representation of change management for versioning can be addressed by reusing some elements from the versioning approach of Workflow4ever, e.g., \textit{roevo:Change}, \textit{roevo:ChangeSpecification} and \textit{roevo:VersionableResource}. \par

Deciding which type of approach should be used for role representation should be based on the needs for either a fine grained definition of the role/relator pattern (a reified relationship), such as the \textit{prov:Association }approach, or a simple property, such as \textit{dc:creator}. While the former (1) enriches the definition of the role (an improved representation capability), the later (2) is less verbose:\par

\begin{lstlisting}[label=lst:lst11, breaklines=true, basicstyle=\small\tt]
(1)
?Association prov:agent opredict:Remzi; 
prov:hadRole opredict:Creator; 
prov:hadPlan ?plan.

(2)
?plan dc:creator 'Remzi' .
\end{lstlisting}

One of the main challenges is to understand the different terminology used for similar conceptualizations. Although the definitions of terms like plan, process, protocol, procedure, workflow, plan specification and SOP seem to be the same (or quite overlapping), it is notorious that when they are used by different communities they carry specific semantics. How to grasp these differences is a crucial question that needs further exploration. For example, in the bioinformatics community, the term \textit{Workflow }usually refer to an implemented (computational) piece of software, i.e., a set of programming language instructions, usually developed with a \textit{Workflow Management System} as a \textit{Workflow Application} \citep{Da-Cruz2012}. On the other hand, in software engineering, the \textit{Workflow }term is usually referred to a detailed business process within the Business Process Modelling (BPM) research. Usually, the BPM languages conform to graphical notations (e.g., BPMN, EPC, ARIS), targeted to human comprehension rather than computational ends (process design/modelling). On the other hand, some BPM languages focus on representing, in a lower-level of abstraction, the process execution details, e.g., BPEL (process implementation) \citep{Rosemann2015} . This is a topic extensively exploited by Service Oriented Architecture (SOA) initiatives. Several related work target the gap that exists between business process models and workflow specifications and implementations, such as service composition schemes \citep{Stephan2012}  and formal provenance of process executions and versioning \citep{Krishna2019}. \par

In\ future work we shall characterize the abstraction levels of workflows based on multi-level process modelling approaches, such as the widespread adopted APQC's Process Classification Framework (PCF). The PCF provides 5 abstraction levels for process specification, from a high abstraction level to detailed workflow specification: category (level 1), process group (level 2), process (level 3), activity (level 4) and task (level 5). Although this framework aims at providing a methodological approach for business process specification, we should investigate whether the minimal information elements of  each level require proper representation in the ontology. We should also consider that challenges of process refinement ($``$process description in a more fine-grained representation$"$ ) \citep{Ren2013} . A process refinement mechanism maps and/or derives models from a higher level specification to a detailed level, equivalent to vertical and exogenous model transformations in model-driven engineering. Typical refinement categories will be investigated, such as activity decomposition principles about event delivering and execution condition transferring \citep{Jiang2016,Muehlen2004}  . The representation of intentionality of the activities within business processes will also be addressed in future work through goal-oriented semantic process modeling \citep{Horkoff2019}, linking goals to activities and roles.\par

Industry-oriented approaches are also being investigated, such as Extract, Transform and Loading (ETL/ELT) for data warehousing, such as SQL Server Integration Services, which considers a workflow as a control flow, while source to destination flows of data are performed by data flows. Furthermore, Product Line Management (PLM) tools should be investigated, especially the ones that cover Laboratory Information Management System (LIMS), which provides important concepts such as Bill-of-Materials (BoM), specifications and their certifications. For example, in PLM a \textit{specification} is a description of raw materials and packaging materials, and semi-finished and finished products [Ref.Siemens]. This description may contain product characteristics (e.g., chemical compounds), recipes (e.g., BoM), production methods, quality norms and methods, artwork, documents, among others.\par
The role of CWL, CEDAR and FAIRsharing should be investigated for improved workflow reproducibility. Ultimately, these initiatives may be used as building blocks for the envisioned FAIR workbench tool, which can be a reference implementation over a workflow system such as Jupyter Notebook (e.g., a plug-in). Finally, the validation of the reproducibility level of a workflow should consider specific FAIR metrics that takes in consideration specific recommendations (e.g., from CWLProv approach) and the practices for higher reproducibility of Jupyter notebooks \citep{pimentel2019large} . \par

\section*{Conclusions}

In this work, we examined how the FAIR principles can be applied to scientific workflows. We adapted the FAIR principles to make the PREDICT workflow, a drug repurposing workflow based on machine learning, open and reproducible. Therefore, the main contribution of this paper is the OpenPREDICT case study, which demonstrates how to make a machine learning workflow FAIR and open. For this, we have created an ontology profile that reuses several semantic models to show how a workflow can be semantically modeled. We published the workflow representation, data and meta-data in a triple store which was used as FAIR data point. In addition, new competency questions have been defined for FAIR workflows and how these questions can be answered through SPARQL queries. Among the main lessons learned, we highlight how the main existing workflow modelling approaches can be reused and enhanced by the profile definition. However, reusing these semantic models showed to be a challenging task, once they present reproducibility issues and different conceptualizations, sometimes overlapping in their terminology. A limitation of this work is that it requires a human-intensive effort to apply the ontology profile on existing workflows, where workflow versioning, prospective and retrospective provenance are manually formalized. To overcome this issue we are developing the FAIR workbench, an implementation reference as a Jupyter Notebook plug-in that facilitates the workflow semantic annotation.

\section*{Acknowledgments}

\bibliography{sample}

\end{document}